\documentclass[a4paper,conference]{IEEEtran}
\usepackage{cite}

\ifCLASSINFOpdf
\else
\fi
\usepackage{amsmath}
\usepackage{amssymb}
\newcommand{\cmark}{\ding{51}}%
\newcommand{\xmark}{\ding{55}}%
\usepackage{pifont}
\usepackage{adjustbox}
\usepackage{subcaption}
\usepackage{xurl}
\usepackage{algorithmic}
\usepackage{algorithm}
\usepackage{graphics}
\usepackage{multirow}
\usepackage{array}

\begin{document}
%
\title{InducT-GCN: Inductive Graph Convolutional Networks for Text Classification}

\author{\IEEEauthorblockN{Kunze Wang, Soyeon Caren Han$^*$, Josiah Poon}
\IEEEauthorblockA{School of Computer Science, The University of Sydney
\\
kwan4418@uni.sydney.edu.au, \{caren.han, josiah.poon\}@sydney.edu.au}
}


%


\maketitle

\let\thefootnote\relax\footnotetext{$*$ Corresponding author (Caren.Han@sydney.edu.au)}

\begin{abstract}
Text classification aims to assign labels to textual units by making use of global information. Recent studies have applied graph neural network (GNN) to capture the global word co-occurrence in a corpus. Existing approaches require that all the nodes (training and test) in a graph are present during training, which are transductive and do not naturally generalise to unseen nodes. To make those models \textit{inductive}, they use extra resources, like pretrained word embedding. However, high-quality resource is not always available and hard to train. Under the extreme settings with no extra resource and limited amount of training set, can we still learn an inductive graph-based text classification model? In this paper, we introduce a novel inductive graph-based text classification framework, InducT-GCN (\textbf{I}nduc\textbf{T}ive \textbf{G}raph \textbf{C}onvolutional \textbf{N}etworks for \textbf{T}ext classification). 
Compared to transductive models that require test documents in training, we construct a graph based on the statistics of training documents only and represent document vectors with a weighted sum of word vectors. We then conduct one-directional GCN propagation during testing. Across five text classification benchmarks, our InducT-GCN outperformed state-of-the-art methods that are either transductive in nature or pre-trained additional resources. We also conducted scalability testing by gradually increasing the data size and revealed that our InducT-GCN can reduce the time and space complexity. The code is available on: \url{https://github.com/usydnlp/InductTGCN}.
\end{abstract}


%
\IEEEpeerreviewmaketitle

\section{Introduction}
Text classification is one of the most fundamental natural language processing research tasks, including topic classification, news categorisation, and sentiment analysis. The aim of the text classification is to classify/categorise textual documents into the predefined classes. The high-dimensional textual input is assigned to the output class with binary or multi-class classification. Note that we only consider a single-label text classification problem in this research. 

Many recent text classification studies have focused on learning text representations using sequence-based learning models, such as convolutional neural networks (CNN) \cite{kim2014convolutional} or recurrent neural networks (RNN) /long short term memory (LSTM) \cite{zhou2016text}. The CNN/RNN-based models focus on the locality and sequence of text and mainly aim to detect semantic and syntactic information in local consecutive word sequences. It tends to neglect global word co-occurrence in a corpus and ignore non-consecutive and long-distance semantic information \cite{peng2018large}. However, those models need a relatively large training set to perform better. Still, most real-world cases (e.g., specific domain or some low resource languages) have a very limited amount of training set (limited labeled data). Recently, pre-trained models, like BERT \cite{devlin2018bert}, RoBERTa \cite{DBLP:journals/corr/abs-1907-11692}, have achieved state-of-the-art performance on several NLP tasks with limited amount of training data. However, those models require much computation and external resources for pre-training, which are not always available.

\cite{yao2019graph} proposed TextGCN and performed well, especially when the percentage of training data is low without using any external resources and with low computation costs. It is an initial text GNN framework, which conducts a straightforward manner of graph construction and applies a GCN-learning\cite{kipf2017semi} to deal with complex structured textual data and prioritise global feature exploitation. More recent studies \cite{wu2019simplifying, liu2020tensor, zhu2021simple} utilise more contextual information or optimising the computation.

However, most graph models are intrinsically transductive. The learned node representations/embeddings for words/documents are not naturally generalisable to unseen words/documents, making it challenging to apply in the real world. The transductive nature of these graph-based learning models requires relatively large computational space when the corpus size is large. Therefore, an inductive model is needed. To expand a transductive graph-based text classification into an inductive model, we mainly consider the following three requirements: 1)The inductive learning model must not include any test set information during the training. 2)The inductive model must not re-train the model on the whole new graph when it learns a new sample. 3)We use corpus-level graph-based text classification to make an inductive model. It nicely covers the benefit of GNN, which captures the complex global structure of the whole corpus and prioritises global feature exploitation.
In this paper, we propose a novel inductive graph-based text classification framework, called \textbf{InducT-GCN}
(\textbf{InducT}ive \textbf{G}raph \textbf{C}onvolutional \textbf{N}etworks for \textbf{T}ext classification). We introduce a new inductive graph framework of graph construction, learning, and testing, and it can expand to any transductive GCN-based text classification model. The paper includes the following contributions:
\begin{itemize}
    \item To the best of our knowledge, we introduce the first inductive corpus-level GCN-based text classification framework without using extra resources.
    
    \item We compare our InducT-GCN on five benchmark datasets under the limited labeled data settings. InducT-GCN outperforms on four of them, even beating some transductive baselines integrated using external resources. 
    
    \item We introduce a new way to make transductive GCN-based text classification models \textit{inductive}, which improves the performance and reduces the time and space complexity.

\end{itemize}

\section{Related Work}

\subsection{Graph Neural Networks}
Graph Neural Network (GNN)s \cite{kipf2017semi} have been effective at tasks to have rich relational structure and can preserve global structure information of a graph in graph embeddings by aggregating first-order neighbourhood information. \cite{kipf2017semi} introduced Graph Convolutional Networks (GCN) on transductive classification tasks. GraphSage \cite{hamilton2017inductive}, and FastGCN \cite{chen2018fastgcn} tailored GCNs on inductive representation learning framework with sampling methods. Graph Attention Networks (GAT)  \cite{velickovic2018graph} applied the Attention to specify different weights to different nodes in a neighbourhood. 
More recent GCN studies for transductive and inductive frameworks have been proposed. For transductive-based GCN, SGC \cite{wu2019simplifying} was introduced to reduce the complexity and S$^2$GC \cite{zhu2021simple} was proposed to solve over-smoothing problems. 
Some inductive-based models, DeepGL \cite{rossi2020deep} and TGAT \cite{tgat_iclr20}, were introduced to cover different graph tasks, including transfer learning and topology learning.

\subsection{Text Classification Using GNN}
GNNs have received attention in various NLP tasks \cite{bastings2017graph, tu2019multi, li2019classifying, yao2019graph,cao2019multi, yang2021hgat}, including text classification. The GNN-based text classification models can be categorised into two types, Document-level and Corpus-level approaches.

\textbf{Document-level GNN in Text Classification}  
Several graph-based text classification models build a graph for each document using words as nodes \cite{defferrard2016convolutional, peng2018large, zhang2018sentence, nikolentzos2020message, huang2019text, zhang2020every}. Word nodes are represented by external resources, pre-trained embedding, such as Word2vec \cite{DBLP:journals/corr/abs-1301-3781}, and Glove \cite{pennington2014glove}. The edges are built either using word co-occurence information\cite{peng2018large,zhang2020every} or simply connecting concetuive words in a sentence\cite{huang2019text}. Hence, they do not consider explicit global structure information of a corpus/entire dataset during their model training/learning.

\textbf{Corpus-level GNN in Text Classification}
TextGCN \cite{yao2019graph} was proposed to build a graph for the entire text corpus with documents and words as nodes. Hence, it captures global information of an entire corpus and conduct node(document) classification. SGC \cite{wu2019simplifying} and S$^2$GC \cite{zhu2021simple} constructed a graph as TextGCN, but proposed different information propagation approaches.
Both TensorGCN\cite{liu2020tensor} and TextGTL\cite{textgtl_ijcai21} proposed three graphs to cover three different aspects. Note that all three graphs are based on an entire corpus and use the same propagation as GCN \cite{kipf2017semi}. TG-Transformer\cite{tgtransformer_emnlp20} applied transformer with pretrained GloVe embeddings to the TextGCN, and BERTGCN\cite{lin2021bertgcn} applied BERT embedding to the TextGCN. All the above models are transductive-based approaches as GCN \cite{kipf2017semi}.  
However, our model InducT-GCN, an inductive graph-based text classification framework, constructs a corpus-level graph but adopts the nature of inductive learning to generalise to unseen nodes naturally.

\section{InducT-GCN}\label{sec:induct}
We propose an Inductive Graph Convolutional Network (GCN) for text classification, named `InducT-GCN', which can be an extension of the traditional transductive GCN-based text classification models. We adopt the traditional transductive GCN-based text classification models, including TextGCN\cite{yao2019graph} and SGC\cite{wu2019simplifying}, and focus on expanding those models to efficient inductive learning models. This section demonstrates the proposed inductive learning components applied to TextGCN.

\subsection{Revisit TextGCN}
TextGCN is a GCN-based text classification model that uses a large  text graph based on the whole corpus. To understand the concept properly, we first explore the GCN process.

\textbf{Graph Convolutional Networks(GCN)}
Formally, considering a graph $G = (V,E,A)$, where $V(|V| = n)$ is a set of nodes, $E$ is a set of edges, and $A \in R^{n\times n}$ is an adjacency matrix representing the edge values between nodes. The propagation rule of each hidden layer is:
\begin{equation}
    H^{(l+1)} = f(H^{(l)}, A) = \sigma (\tilde{A}H^{(l)}W^{(l)})
\end{equation}
where $\tilde{A}=D^{-\frac{1}{2}}AD^{-\frac{1}{2}}$ is a normalized symmetric matrix for $A$ and $D_{ii }= \Sigma_jA_{ij}$ as a degree matrix of adjacency matrix $A$. $H^{(l)}$ is $l_{th}$ hidden layer input and $W^{(l)}$ is the weight to be learned in this layer. $\sigma$ is an activation function, e.g. ReLU: $\sigma(x) = max(0,x)$.

\textbf{TextGCN}
Followed by the GCN\cite{kipf2017semi}, TextGCN constructs a large corpus-level graph but with textual information, documents and words as nodes so it can model the global word-document co-occurrence. The constructed graph includes documents and words nodes from training sets and test sets. TextGCN aims to model the global word-document occurrence with two major edges: 1) word-word edge: calculated by co-occurrence information point-wise mutual information(PMI)\cite{aji2012document}, 2) document-word edge: TF-IDF. One-hot vectors are fed into a two-layer GCN model to jointly learn the embedding for the documents and words. The representations on the document nodes in the training set train the classification model while those in the test set are used for prediction.

\subsection{Tranductive and Inductive Nature}\label{sec:trans}
This section discusses the nature of transductive and inductive GCN learning for text classification and what inductive learning aspect we would like to explore. Most GCN text classification models, including TextGCN\cite{yao2019graph},  SGC\cite{wu2019simplifying}, or S$^2$GC \cite{zhu2021simple}, are inherently transductive by using the whole corpus, including training set and test set all time. 
 
To expand those transductive models into an inductive learning nature, we fundamentally improve two aspects as follows. First, the transductive GCN-based text classification models include documents from the training set and the test set when constructing a whole-corpus-based textual graph for GCN learning. Hence, the learned GCN model will be influenced/generalised by word/document information in the test set, which is supposed to be unseen nodes. Our inductive GCN-based text classification model constructs a graph with only training document information but does not consider any information from the test sets. We focus on generalising to unseen nodes and aligning newly observed subgraphs to the node that the model has already optimised on.  

Secondly, the transductive models learn the embedding for $V_{train}$,  $V_{test}$,  $V_{word}$  simultaneously by using one-hot input vectors $H^{(0)} \in R^{n \times n}$. For any new test sample, the embedding should be re-learned by re-training the model on the new graph. In this case, the re-learning/re-training process does not perfectly fulfil the effective generalisation to unseen nodes. Therefore, we develop a new graph construction and training/testing solution for inductive learning instead of re-learning or re-training.

\subsection{InducT-GCN Graph Construction}
\subsubsection{Graph Nodes}
Our inductive GCN-based text classification model, InducT-GCN, strictly do not consider any information or statistics from the test set, which is supposed to be unseen nodes. Instead, we construct the nodes only with training document information. Consider a set of nodes $V = \{V_{train},V_{word}\}$ and the $V_{word}$ are the unique words in the training documents. To define input vector $H^{(0)}$ for graph nodes in the InducT-GCN graph, we consider two requirements:
(1)During the propagation phase, the graph is considered as a homogeneous graph, which means all the nodes are considered as the same type without checking whether they are word nodes or document nodes. Then all the input vectors for document nodes and word nodes should align with each other. 
(2)Our InducT-GCN must not use one-hot vectors for representing document nodes to avoid learning any representation on testing documents during training.

\begin{figure}[t]
    \centering
    \includegraphics[scale=0.55]{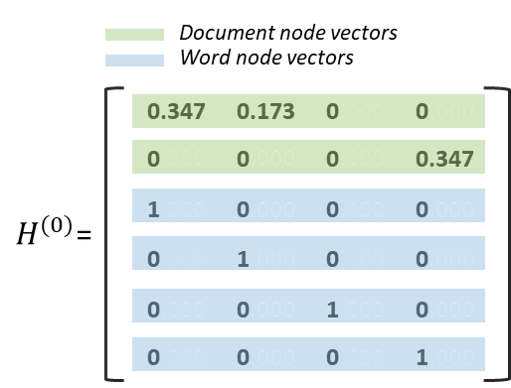}
    \caption{Input Vectors Representations when two input documents are ``$word_1$ $word_1$ $word_2$ $word_3$" and ``$word_3$ $word_4$".}
    \label{fig:input}
\end{figure}

With this in mind, we propose a new document representation by focusing on the nature of our proposed inductive learning idea. InducT-GCN generates document node representations based on its word node vectors for the proper alignment between word and document representation. We use a weighted average of word vectors to construct document nodes vectors, and the key idea of this construction is applying TF-IDF weights. Formally, one-hot vectors are used for representing word nodes vectors $H^{(0)}_i, \forall i \in V_{word}$. For representing training documents node vectors $H^{(0)}_i, \forall i \in V_{train}$, we use TF-IDF vectors.
The values for each dimension is TF-IDF values for the corresponding word in that specific document:
$H^{(0)}_{ij} = \text{TF-IDF}(i,j)$
where $i$ and $j$ are document and word, respectively. Figure \ref{fig:input} shows an example of $H^{(0)}$.

\subsubsection{Graph Edges}
We focus on expanding corpus-level transductive GCN-based text classification to inductive learning and select TextGCN\cite{yao2019graph} as one of its kind. Like TextGCN, we define two-edge types for the InducT-GCN graph: 1) Word-Word with PMI and 2) Word-Doc edges with TF-IDF. Note that each node also connects to itself. PMI is calculated based on the co-occurrence of a pair of words in a slicing window. Formally, it is calculated by:
$\text{PMI}(i,j) = log \frac{p(i,j)}{p(i)p(j)}$
where $p(i,j)$ represents the co-occurrence probability of word $i$ and $j$ and estimated by $p(i,j) = \frac{\# Co-occurrence}{\# Windows}$, $p(i)$ represents the probability of word $i$ and estimated by $p(i) = \frac{\# Occurrence}{\# Windows}$. The graph is un-directed and all the edges are symmetrical. 

\subsection{InducT-GCN Learning and Testing}
\label{sec:learning}

After building the graph, we train it using a two-layer GCN as in \cite{kipf2017semi}. The first GCN layer learns the word embeddings. The dimension of the second GCN layer is the number of classes of the dataset, and the output is fed into a softmax activation function. For example, a binary classification task will result in imension of the second GCN layer as 2. The node representations on the training documents are used for cross-entropy loss calculation and back-propagation. 
Formally, the propagation can be described as:
\begin{gather}
    H^{(1)} = \sigma (\tilde{A}H^{(0)}W^{(0)}) \\
    Z = softmax(\tilde{A}H^{(1)}W^{(1)})
\end{gather}
where $W^{(0)}$ is a learned word embedding lookup table, and  $W^{(1)}$ is learned weight matrix in second layer. Loss is calculated by using cross-entropy between $Z_i$ and $Y_i$, $\forall i \in V_{train}$.

In GCN, the propagation for each layer is conducted by updating the nodes with the weighted sum of their first-order neighbours and the node itself. In order to make predictions on the test set, the first-order and second-order neighbours' representation for each test document should be aggregated. Note that we utilise the test documents during the testing phase only, so there is no need to update all the nodes in the graph during the propagation. 

Instead, we conduct an one-direction propagation and only update test document nodes. Firstly, $H^{(1)}_i, \forall i \in V_{word}$, $W^{(0)}$ and $W^{(1)}$ are recorded after training phase, and $H^{(1)}_i, \forall i \in V_{word}$ is notated as $H^{(1)}_{word}$.
Storing $H^{(1)}_{word}$ enables InducT-GCN not to work on the training document nodes during the test phase, so it saves computation resources. During the testing phase, InducT-GCN supports batch testing\cite{hamilton2017inductive}. For each batch of test document node $B \in V_{test}, |B|=b$, the edges $E_{B}$ between $B$ and $V_{word}$ are calculated using TF-IDF with the document frequency of the training set.

\begin{algorithm}[t!]
\caption{InducT-GCN Training and Testing Phase}
\label{alg:induct1}
\textbf{Input}: Training Graph $G(V,\tilde{A}), V = \{V_{train},V_{word}\}$;\\
Training input vectors $H^{(0)}$;\\
Training Labels \{$Y_{i}, \forall i \in V_{train}\}$;\\
Adjacency Matrix for Test Batch Subgraph\{$A_B,B \in V_{test}$\}; \\ Test input vectors $\{H'^{(0)}_{B}, B \in V_{test}\}$ \\
\textbf{Parameter}: Weight matrices $W^{(0)}$ and $W^{(1)}$ \\
\textbf{Output}: \text{Prediction Results} \{$Y_{B}, \forall B \in V_{test}\}$\\
\vspace{-5mm}
\begin{algorithmic}[1]
\FOR {$epoch=1,2,\ldots$}
	\STATE $H^{(1)} \gets \sigma (\tilde{A}H^{(0)}W^{(0)})$
	\STATE $Z \gets softmax(\tilde{A}H^{(1)}W^{(1)})$
	\STATE $L \gets \text{CrossEntropy}(Y_{i},Z_{i}), \forall i \in V_{train}$
	\STATE Update $W^{(0)}$ and $W^{(1)}$
\ENDFOR
\STATE $H^{(1)}_{word} \gets H^{(1)}_i, \forall i \in V_{word}$
\FOR {Batch $B$ in $V_{test}$}
    \STATE $G' \gets \{A_B, H^{(0)}_{word}, H'^{(0)}_B\}$
    \STATE $H'^{(1)}_{B} \gets \text{GCN}(G', W^{(0)})$
    \STATE $G'' \gets \{A_B, H^{(1)}_{word}, H'^{(1)}_{B}\}$
    \STATE $Y_{B} \gets argmax(\text{GCN}(G'', W^{(1)}))$
\ENDFOR
\end{algorithmic}
\end{algorithm}

Test document input $H'^{(0)}_B$ is also calculated using TF-IDF. The testing phase can be described as:
\begin{gather}
    A_B = concat(E_{B},I) \\
    H'^{(0)}_{word,B} = concat(H^{(0)}_{word},H'^{(0)}_B) \\
    H'^{(1)}_B = \sigma (A_BH'^{(0)}_{word,B}W^{(0)}) \\
    H'^{(1)}_{word,B} = concat(H^{(1)}_{word},H'^{(1)}_B) \\
    Z_B = softmax(A_BH'^{(1)}_{word,B}W^{(1)}) \\
    Y_B = argmax(Z_B)
\end{gather}
where $A_B \in R^{b \times (|V_{word}|+b)}$ stands for the weights of the weighted sum calculation and it can be considered as an adjacency matrix for test batch subgraph. $H'^{(0)}_{B} \in R^{(|V_{word}|+b) \times |V_{word}|}$ stands for the test batch input vectors in the subgraph. $H'^{(1)}_B \in R^{b \times h}$ is the updated test document node embedding after the first layer of GCN, and $h$ is the hidden dimension size. Then, the stored $H^{(1)}_{word}$ on the word nodes, which contains the first layer training documents information, are used to propagate training documents information to the test document nodes in the second layer. We formally describe the overall algorithms for the training and testing phase of InducT-GCN in Algorithm \ref{alg:induct1}.


\subsection{Space and Time Analysis}
Compared with TextGCN, InducT-GCN is more efficient both in time and space. For the space complexity:(1)Number of Parameters of InducT-GCN is $|V_{word}|*h+h*c$ while TextGCN requires $(|V_{train}| +|V_{word}|+|V_{test}|)*h+h*c$ parameters. Meanwhile, $|V_{word}|$ in InducT-GCN is smaller than that in TextGCN.
(2)Graph Space complexity of InducT-GCN is $O(|V_{word}|^2+|V_{train}|*|V_{word}|)$ and for TextGCN, it is $O(|V_{word}|^2+(|V_{train}|+|V_{test}|)*|V_{word}|)$. Similarly, $|V_{word}|$ in InducT-GCN is smaller than TextGCN. Compared with TextGCN, our InducT-GCN is faster in three ways:
(1)When constructing the graph, the time complexity of PMI is $O(|V_{word}|^2*\# Windows)$, and it is smaller for InducT-GCN.
(2)The training time is shorter for InducT-GCN since the graph is smaller.
(3)When testing on new samples, TextGCN requires retraining while InducT-GCN can make predictions without retraining.

\begin{table}[t!]
\centering
\begin{adjustbox}{width=1\linewidth}
\begin{tabular}{c|ccccccc}
\hline
\textbf{Dataset} & \textbf{\# Train} & \textbf{\# Test} & \textbf{\# Word} & \textbf{\# Class}& \textbf{Avg Len}\\
\hline
R8 & 274 & 2,189 & 1,878  & 8 & 62.22\\
R52 & 326 & 2,568 & 2,568  & 52 & 66.98\\
Ohsumed & 167 & 4,043 & 2,667  & 23 & 123.7\\
20NG & 113 & 7,532 & 2,839  & 20 & 163.5\\
MR & 355 & 3,554 & 605  & 2 & 7.25\\
\hline
\end{tabular}
\end{adjustbox}
\caption{\label{tab:dataset} Summary statistics of datasets.}
\end{table}

\begin{table*}[h!]
\centering
\begin{adjustbox}{width=0.95\linewidth}
\begin{tabular}{c|c|c|c|c|c|c}
\hline
\textbf{Method}  & \textbf{PT.} & \textbf{R8 5\%} & \textbf{R52 5\%} & \textbf{Ohsumed 5\%} & \textbf{20NG 1\%} & \textbf{MR 5\%} \\
\hline
TF-IDF + SVM  & \xmark & 0.8054 ± 0.0000 & 0.6830 ± 0.0000 & 0.1476 ± 0.0000 & 0.1289 ± 0.0000 & 0.5537 ± 0.0000 \\
TFIDF + LR & \xmark & 0.8090 ± 0.0000 & 0.6869± 0.0000 & 0.1813 ± 0.0000 & 0.1860 ± 0.0000 & 0.5967 ± 0.0000 \\\hline
CNN-rand  & \xmark & 0.8107 ± 0.0110 & 0.6854 ± 0.0100 & 0.1586 ± 0.0079 & 0.1390 ± 0.0179 & 0.5485 ± 0.0122 \\
CNN (Pretrain)  & \cmark  & 0.9052 ± 0.0097 & 0.7708 ± 0.0181 & 0.3411 ± 0.0370 & 0.2969 ± 0.0277 & \textbf{0.7009 ± 0.0060} \\
LSTM-rand  & \xmark & 0.7392 ± 0.0146 & 0.6364 ± 0.0060 & 0.1614 ± 0.0085 & 0.0766 ± 0.0063 & 0.5301 ± 0.0191 \\
LSTM (Pretrain)&  \cmark & 0.7916 ± 0.0499 & 0.6667 ± 0.0303 & 0.2486 ± 0.0392 & 0.1010 ± 0.0220 & 0.6680 ± 0.0198 \\\hline
TextGCN \cite{yao2019graph} & \xmark & 0.9116 ± 0.0127 & 0.7885 ± 0.0751 & 0.2225 ± 0.1138 & 0.2198 ± 0.1293 & 0.5341 ± 0.0216 \\
SGC \cite{wu2019simplifying} & \xmark & 0.8955 ± 0.0098 & 0.7725 ± 0.0189 & 0.2474 ± 0.0392 & 0.2948 ± 0.0342 & 0.6015 ± 0.0051 \\
TextING \cite{zhang2020every} &  \cmark & 0.8648 ± 0.0414 & 0.7465 ± 0.0298 & 0.3026 ± 0.0235 & N/A & 0.6117 ± 0.0342 \\\hline
InducT-SGC& \xmark & 0.9045 ± 0.0046 & 0.8046 ± 0.0066 & 0.3106 ± 0.0061 & 0.2990 ± 0.0251 & 0.6017 ± 0.0048\\
InducT-GCN & \xmark & \textbf{0.9155 ± 0.0051} & \textbf{0.8135 ± 0.0384} & \textbf{0.3563 ± 0.0078} & \textbf{0.3461 ± 0.0337} & 0.6037 ± 0.0038 \\
\hline
\end{tabular}
\end{adjustbox}
\caption{\label{tab:baseline1} Comparison of with Baseline on Limited Labeled Data. For 20NG dataset, TextING\cite{zhang2020every} has out of Memory issue and they also have not tested on the 20NG either. *The column PT. refers to the model applied any pre-trained embedding.}
\end{table*}


\section{Evaluation Setup}
We evaluate the performance of our InducT-GCN on text classification and examine the effectiveness of the proposed inductive learning approach.

\subsection{Dataset}\label{sec:dataset}
We first evaluate InducT-GCN on 5(five) publicly available text classification benchmark datasets, including R8, R52, Ohsumed, 20NG, and MR. To test in the limited labelled data environment, we select 5\% of the full training set (1\% for 20NG, due to the size) and remain the original test size. The detailed statistics of datasets can be found in Table \ref{tab:dataset}. We also apply the data augmentation methods on R8 test set to evaluate on the larger test sets, called R8A. 
\textbf{R8}, \textbf{R52} are two subsets of the Reuters dataset and focus on the topic classification. \textbf{Ohsumed} is produced by the MEDLINE database, containing cardiovascular diseases abstracts.  \textbf{20NG}(20 NewsGroup) is a 20 class-based news classification dataset. 
\textbf{MR}(Movie Review) is a binary (positive and negative) sentiment polarity analysis. 
\textbf{R8A}: To evaluate our InducT-GCN scalability in the larger test set, we apply a data augmentation technique. 
    Nlpaug\cite{ma2019nlpaug} is applied for augmenting the R8 test set. To achieve this, we randomly choose one of the four options: (1) randomly deleting a word, (2) adding a word based on Word2Vec embedding similarity, (3) substituting a word using Synonyms in WordNet \cite{miller1995wordnet}, (4) randomly swapping two words. The detailed evaluation can be found in Section  \ref{sec:test_size_testing}.
 
 All datasets are preprocessed based on \cite{kim2014convolutional}. We remove the words if shown less than twice in the training documents since words only shown once can not work as a bridge between two document nodes. Words listed in the NLTK stopwords list are alsoe removed. We apply the same preprocessing method for all experiments.

\subsection{Baselines}
We compare InducT-GCN with baselines, mainly those models with no external resources and learning inductively. However, due to the limited number of baselines, we include the baselines with pre-trained word embeddings, such as CNN (Pretrain), LSTM (Pretrain), and TextING. We also add transductive models, including TextGCN and SGC.

\begin{itemize}
    \item \textbf{TF-IDF + SVM/LR} applies TF-IDF vectors and uses Support Vector Machine (SVM) or Logistic Regression (LR) as classifiers respectively.
    
    \item \textbf{CNN}/\textbf{LSTM}\cite{kim2014convolutional,DBLP:conf/ijcai/LiuQH16} apply CNN and Long Short-Term Memory with randomly initialised word embeddings or pretrained GloVe\cite{pennington2014glove} embeddings.

    \item \textbf{TextGCN/SGC/TextING} \cite{yao2019graph,wu2019simplifying,zhang2020every} are GNN text classification models. TextGCN and SGC are corpus-level GNN and TextING is document-level GNN.
\end{itemize}

\subsection{Settings}
We apply the same set of hyper-parameters to all datasets without hyper-parameter tuning for a fair comparison. For TextGCN \cite{yao2019graph}, SGC \cite{wu2019simplifying}, our InducT-GCN and InducT-SGC, as described in \cite{yao2019graph}, we applied two layers graph convolutional, and the hidden dimension is 200. Adam optimizer with a learning rate of 0.02 is used for training. For each experiment, followed by \cite{kipf2017semi}, 200 epochs are set to be the maximum number of epochs, and early stopping of 10 epochs is applied. 10\% of the training set is randomly selected as the validation set. An early stopping mechanism is also applied for other baseline models by using the default hyperparameters. 

Followed by \cite{yao2019graph,wu2019simplifying, zhang2020every}, we use the accuracy as an evaluation metric and produce the average and standard deviation of the ten-time running results for each testing result.

\begin{table*}
\centering
\begin{adjustbox}{width=0.9\linewidth}
\begin{tabular}{c|c|c|c|c|c}
\hline
\textbf{Embedding} & \textbf{R8 5\%} & \textbf{R52 5\%} & \textbf{Ohsumed 5\%} & \textbf{20NG 1\%} & \textbf{MR 5\%}\\
\hline
Random & 0.9155 ± 0.0051 & 0.8135 ± 0.0384 & \textbf{0.3563 ± 0.0078} & 0.3461 ± 0.0337 & 0.6037 ± 0.0038\\
Pretrained Word2Vec & 0.9124 ± 0.0043 & \textbf{0.8290 ± 0.0084} & 0.3544 ± 0.0305 & 0.3476 ± 0.0086 & 0.6003 ± 0.0045\\
Pretraind GloVe & \textbf{0.9159 ± 0.0102} & 0.8266 ± 0.0090 & 0.3514 ± 0.0186 & \textbf{0.3662 ± 0.0197} & \textbf{0.6051 ± 0.0055}\\
\hline
\end{tabular}
\end{adjustbox}
\caption{\label{tab:emb}Test Accuracy with Different Initialized Embedding Method}
\end{table*}

\section{Result}
\subsection{Performance Evaluation}\label{sec:gene_eval}
A comprehensive experiment is conducted on the five benchmark datasets in the limited environment as mentioned in Section \ref{sec:dataset}. The result presented in Table \ref{tab:baseline1} shows that our proposed InducT-GCN significantly outperforms all baselines in terms of average accuracy on four datasets in R8, R52, Ohsumed, 20NG. Note that the baselines include transductive graph-based models, such as TextGCN and SGC, and CNN, LSTM, TextING use external resources, like pre-trained word embeddings. Meanwhile, the standard derivation is smaller than most baseline models, showing the robustness of our model.

For more in-depth performance analysis comparing our InducT-GCN with baseline models, we can highlight that this result shows the effectiveness of the proposed InducT-GCN on long text datasets. While the average lengths of 20NG and Ohsumed are longer than 100 and those of R8 and R52 are still longer than 60, MR is less than 10, which can be considered as an extremely short text dataset. We found that models with pre-trained word embeddings GloVe perform better on short text documents. This is mainly because it would be challenging to recognise the global word co-occurrence with this short text document length, leading to fewer connections (bridging) between word nodes in corpus-level text graphs. Nevertheless, our InducT-GCN performs the best among the models with no pre-trained embeddings.

With our Inductive graph construction and learning framework, it is possible to expand to any corpus-level and transductive GCN-based text classification models, such as TextGCN\cite{yao2019graph}, SGC\cite{wu2019simplifying}, TensorGCN\cite{liu2020tensor}, and S$^2$GC \cite{zhu2021simple}. This section reports the generalisation capability of our inductive graph construction and learning framework. We now expand our inductive framework to another corpus-level graph-based model, SGC \cite{wu2019simplifying}, and called InducT-SGC. 
Table \ref{tab:baseline1} also visualises the comparison of the original transductive SGC models and our InducT-SGC. As shown in the table, our InducT-SGC produces much higher performance than the original SGC when the labeled data are limited. The performance improvement between both pairs of the original transductive and our inductive model, TextGCN-to-InducT-GCN and SGC-to-InductSGC, clearly shows the generalisation capability of our proposed inductive framework. It can also be applied to other corpus-level graph-based text classification models in the future. 

\begin{figure} [t]    
     \begin{subfigure}{0.49\linewidth}
         \includegraphics[width=\linewidth]{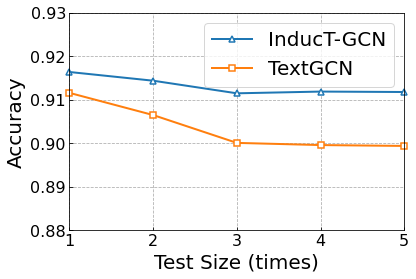}
         \caption{TextGCN}
         \label{fig:testsizea}
     \end{subfigure}
     \begin{subfigure}{0.49\linewidth}
         \includegraphics[width=\linewidth]{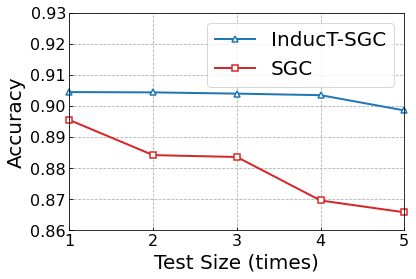}
         \caption{SGC}
         \label{fig:testsizeb}
     \end{subfigure}
     \caption{Test accuracy with different test size on R8A by using TextGCN, SGC and our proposed Inductive model.}
     \label{fig:testsize}
\end{figure}

\subsection{Impact of Test Size}\label{sec:test_size_testing}
As mentioned earlier, we use the R8A (R8 with a data augmentation) to show the scalability of our proposed Inductive learning framework by comparing TextGCN and InducT-GCN in the larger text set. 
Figure \ref{fig:testsizea} shows the comparison of TextGCN and InducT-GCN on different test sizes, with 1 to 5 times (2,189, 4,378, 6,567, 8,756, 10,945) of the R8 original test set size. The larger the test size, the larger the gap between TextGCN and InducT-GCN. TextGCN produces worse performance with the largest test size. This is mainly because only a small proportion of the document nodes would contribute to the gradient in TextGCN with a larger test set. Especially during the training phase, it is difficult for TextGCN to learn embeddings of those test document nodes having fewer connections with word nodes by backpropagation. 
Moreover, we found that the performance of our InducT-GCN decreased only a little (less than 0.5) when the test size grew. We also conducted the same evaluation based on SGC by applying our Inductive graph construction and learning framework to SGC, InducT-SGC. Like the result that our InducT-GCN produced, the InducT-SGC delivers a much higher performance than the original SGC. The performance trend shows how our Inductive framework perfectly fits the inductive learning nature.  

\subsection{Impact of Initial Word Embedding}
As mentioned in Section \ref{sec:learning}, the first layer of InducT-GCN learns the word embeddings and is randomly initialized. We also examine other initial embedding weights methods including pre-trained Word2vec \cite{DBLP:journals/corr/abs-1301-3781} and GloVe \cite{pennington2014glove}. 
Table \ref{tab:emb} shows the performance comparison of different pre-trained word embeddings. In most cases except Ohsumed, pre-trained word embeddings can improve the performance of InducT-GCN. However, Ohsumed is a medical-related dataset, and out-of-vocabulary issue of the pretrained embedding doesn't help on the document classification task.
Although we only focus on the model without using any external resources in this study, this result still shows the potentiality of InducT-GCN when used with external resources.

\subsection{Computation Time Results}\label{sec:time_result}
Table \ref{tab:time} compares the original transductive TextGCN with InducT-GCN on R8A and visualises the superiority of our inductive learning framework by reducing the computation time, including graph construction and training. The larger the test size is, the more time InducT-GCN can save.

\begin{table}[t!]
\centering
\begin{adjustbox}{width=0.9\linewidth}
\begin{tabular}{c|c|c|c|c}
\hline
\multirow{2}{*}{\textbf{Test Size}} & \multicolumn{2}{c|}{\textbf{TextGCN}} & \multicolumn{2}{c}{\textbf{InducT-GCN}} \\ \cline{2-5} 
                                    & \textbf{Graph}   & \textbf{Training}  & \textbf{Graph}    & \textbf{Training}    \\ \hline
$\times$1                                  & 6.29             & 2.75               & 0.89              & 0.52                 \\ \hline
$\times$2                                  & 11.90            & 3.20               & 1.11              & 0.53                 \\ \hline
$\times$3                                  & 16.60            & 3.54               & 1.30              & 0.53                 \\ \hline
$\times$4                                  & 21.10            & 4.13               & 1.52              & 0.55                 \\ \hline
$\times$5                                  & 27.50            & 4.98               & 1.68              & 0.51                 \\ \hline
\end{tabular}
\end{adjustbox}
\caption{\label{tab:time} Graph Construction Time and Training Time comparison on R8A (sec). Hardware: 16 Intel(R) Core(TM) i9-9900X CPU @ 3.50GHz and NVIDIA Titan RTX 24GB}
\end{table}

\begin{table}[t!]
\centering
\begin{adjustbox}{width=0.95\linewidth}
\begin{tabular}{c|c|c}
\hline
\textbf{Method} & \textbf{R8 Full} & \textbf{R52 Full} \\
\hline
TextGCN & 0.9629 ± 0.0010 & 0.9295 ± 0.0020 \\
InducT-GCN & \textbf{0.9653 ± 0.0017} & \textbf{0.9323 ± 0.0015} \\
\hline
\end{tabular}
\end{adjustbox}
\caption{\label{tab:baseline2} Test Accuracy on Full Data Setting}
\end{table}

\subsection{Performance in Full Dataset}\label{sec:full}
We also evaluated the performance of our InducT-GCN with the full dataset, like the TextGCN was evaluated \cite{yao2019graph}. As can be seen in Table \ref{tab:baseline2}, the performance of InducT-GCN and TextGCN on R8 and R52 are comparable when using the entire dataset with the same hyperparamters. We can conclude that InducT-GCN is superior to the TextGCN in terms of performance and computation, which is not only in smaller space with fewer parameter numbers but also in the whole dataset setting.

\section{Conclusion}
This study proposes a novel inductive graph-based text classification framework, InducT-GCN, which makes heavy and transductive GCN-based text classification models inductive. We construct a graph only using training set statistics. InducT-GCN can efficiently capture global information with fewer parameters and smaller space complexity. Our InducT-GCN significantly outperformed all graph-based text classification baselines and was even better than the models using pretrained embeddings. We also demonstrated the generalisation capability of our inductive graph construction and learning framework by applying and expanding different transductive graph-based text classification models, like TextGCN and SGC. Compared to the original models, the performance and computation time were surprisingly improved. It is hoped that this paper provides some insight into the future integration of the lighter and faster inductive graph neural networks on different NLP tasks.

\bibliographystyle{IEEEtran}
\bibliography{IEEEabrv,anthology}
%




\end{document}